\documentclass[10pt,twocolumn,letterpaper]{article}

\usepackage[pagenumbers]{wacv} 

\usepackage{graphicx}
\usepackage{amsmath}
\usepackage{amssymb}
\usepackage{booktabs}
\usepackage{amsthm}
\newtheorem{definition}{Definition}
\newtheorem{theorem}{Theorem}
\newtheorem{corollary}{Corollary}[theorem]
\usepackage{algorithm}
\usepackage{algpseudocode}

\usepackage[pagebackref,breaklinks,colorlinks]{hyperref}

\usepackage[capitalize]{cleveref}
\crefname{section}{Sec.}{Secs.}
\Crefname{section}{Section}{Sections}
\Crefname{table}{Table}{Tables}
\crefname{table}{Tab.}{Tabs.}

\def\wacvPaperID{2408} 
\def\confName{WACV}
\def\confYear{2025}

\begin{document}

\title{Delta-NAS: Difference of Architecture Encoding for Predictor-based Evolutionary Neural Architecture Search}

\author{Arjun Sridhar\\
Duke University\\
Durham, NC\\
{\tt\small arjun.sridhar@duke.edu}
\and
Yiran Chen\\
Duke University\\
Durham, NC\\
}
\maketitle

\begin{abstract}
  Neural Architecture Search (NAS) continues to serve a key roll in the design and development of neural networks for task specific deployment. Modern NAS techniques struggle to deal with ever increasing search space complexity and compute cost constraints. Existing approaches can be categorized into two buckets: fine-grained computational expensive NAS and coarse-grained low cost NAS. Our objective is to craft an algorithm with the capability to perform fine-grain NAS at a low cost. We propose projecting the problem to a lower dimensional space through predicting the difference in accuracy of a pair of similar networks. This paradigm shift allows for reducing computational complexity from exponential down to linear with respect to the size of the search space. We present a strong mathematical foundation for our algorithm in addition to extensive experimental results across a host of common NAS Benchmarks. Our methods significantly out performs existing works achieving better performance coupled with a significantly higher sample efficiency. 



\end{abstract}

\section{Introduction}
\label{sec:intro}
Neural Architecture Search (NAS) \cite{cai2019ofa,liu2019darts,tan2018mnas} enables state-of-the-art performance for a specific task with predetermined hardware and latency constraints. This paradigm is popular in many modern application driven deployments of neural networks such as image recognition, speech detection or on-device large language models (LLMs) where latency, accuracy, and hardware constraints play a large part in the design of neural architectures. As such, NAS serves a vital role in improving machine learning (ML) performance \cite{nas-surevey}. 

NAS algorithms are tasked with determining the best neural architecture within a search space: set of design choices usually defined by a set of operation choices and their connections \cite{lissnas}. Existing NAS methods accomplish the goal of searching through a variety of methods. Early algorithms employ RL-based controllers to design neural architectures. However, the training and search process involved substantial computational costs. Consequently, researchers explored various alternative approaches, such as genetic algorithms \cite{evolv_ss}, Bayesian optimization \cite{white2020bananas}, predictors \cite{strongernasweakerpred}, and one-shot NAS \cite{guo2019singlepath} to tackle these computational complexities. Fundamentally all of these techniques rely on obtaining the accuracy of a given candidate architecture. Accurately learning the relationship between architecture and accuracy is a challenging task as the search space grows. 

More recently, significant strides have been taken by Zeroshot \cite{zeroshot} to reduce the cost associated with training through using alternative methods to predict an architectures accuracy. Zeroshot NAS uses proxy metrics dubed Zen-Score to predict the accuracy of a network without training by estimating the expressivity of the architecture. This technique trades accuracy for compute cost. In contrast, efforts by Fewshot \cite{fewshot} serve to improve the accuracy of NAS with increased training cost and complexity through employing a sharded super network. Both of these methods struggle to preform well on modern larger search spaces.

In recent years, NAS algorithms are tasked with increasingly large search spaces while minimizing the amount of compute resources used for search. Large and complex search spaces have an increased likelihood of containing better-performing candidates. Additionally, technical advancements in operation choices, architecture complexity, and network depth/width have resulted in an increased need for such vast spaces. However, these large and complex search spaces come with a slew of optimization and efficiency challenges. Coupled with the continual desire to cut compute costs, NAS algorithms face challenges of ever-escalating complexity. As the task complexity and demand increases, there has been in increase in the need for efficient and clever algorithms in the domain. 

\begin{figure}[t]
    \centering
    \includegraphics[scale=0.7]{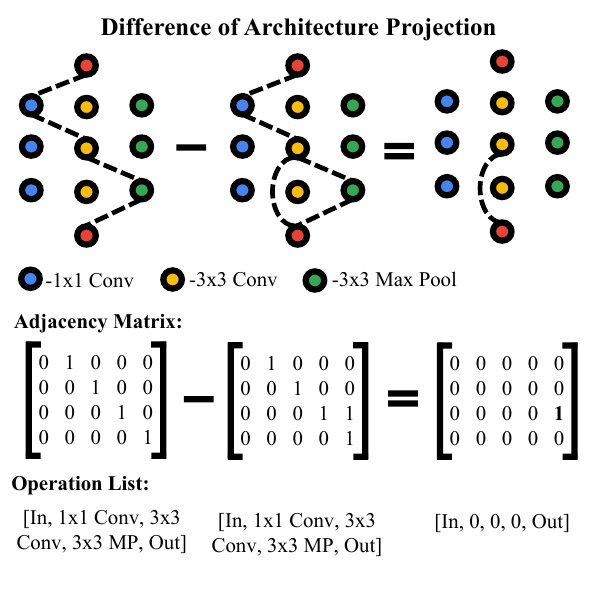}
    \caption{Through taking the difference of architectures that are similar, we are able to project to a sparse representation. We are able to scale to large search spaces linearly.}
    \label{fig:overview}
\end{figure}
We propose a paradigm shift from existing methods through projecting the problem to a lower dimensional space. Instead of mapping networks to their accuracy, we propose mapping a pair of close networks to their difference in accuracy. By taking a difference in networks, Delta-NAS creates a sparse encoding of dense neural candidates as illustrated in Figure \ref{fig:overview}. Through taking the difference of close architectures, Delta-NAS is able to map exponential growth in the input search space to linear growth in the difference of architecture space. This reduction greatly improves NAS performance.

Typically, lossy encoding schemes have been detrimental to NAS performance. Delta-NAS, however, enables a lossy projection while preserving NAS performance through the unique change in target. By predicting differences instead of raw performance, Delta-NAS learns the impact of a change in the architecture. We can then exploit the predictor to evaluate changes to a given architecture to generate promising candidate architectures as a series of improvements.


The main contributions of this work can be summarized as follows:
\begin{enumerate}
    \item We present a theoretical foundation for analysing difference of architecture encoding schemes across common NAS search space types.
    \item To the best of our knowledge, we present the first NAS algorithm that scales linearly to an exponential increase in the number of candidate architectures being considered.
    \item We fill a large void left by current NAS approaches by creating an algorithm which balances compute cost and accuracy tuned specifically for the modern and future NAS landscape. 
\end{enumerate}
Our methods achieves state-of-the-art results across a wide range of NAS benchmarks covering a diverse range of search spaces sizes and ML tasks.

\section{Related Work}
\label{sec:related}
NAS consists of two main parts: an evalutation strategy to obtain the performance of an architecture and a search strategy to obtain the next architecture or set of architectures to evaluate. 

\subsection{Evaluation Strategy: Predictor-based NAS}
Predictor-based NAS utilizes an accuracy predictor to guide the sampling of novel architectures \cite{white2021powerfulperformancepredictorsneural}. BONAS \cite{shi2020bridginggapsamplebasedoneshot} TA-GATES \cite{tagates}, and BRP-NAS \cite{dudziak2021brpnaspredictionbasednasusing} employ predictor-based NAS focusing on sample efficiency. BONAS \cite{shi2020bridginggapsamplebasedoneshot} utilizes a Graph Convoulional Network for accuracy prediction as a surrogate function of Bayesian Optimization. BRP-NAS \cite{dudziak2021brpnaspredictionbasednasusing} employs an iterative sampling strategy combined with a binary relation predictor. Most recently, TA-GATES \cite{tagates} employs learnable encoding for operation choices. While all of these works utilize predictor-based NAS, they use different search strategies to perform the final stage of NAS. 

\subsection{Search Strategy: Evolutionary Methods}
Population-based NAS emulates natural evolution by maintaining a diverse population of architectures and evolving them through mutation and crossover. Mutation allows for local fine-tuning, while crossover drives a more focused global search, serving as the core mechanism of evolutionary discovery. However, most modern evolutionary NAS techniques rely solely on mutation \cite{real2017largescale,real2019regularized,so2022primersearchingefficienttransformers,coreyes2022evolvingreinforcementlearningalgorithms}. The permutation problem arises from isomorphism in the graph space, where functionally identical architectures are represented by different encodings, making crossover operations problematic. Although several solutions have been proposed, none have successfully generalized across different search space types. To address this limitation, Shortest Edit Paths (SEP)\cite{pmlr-v202-qiu23b-sep} introduced a novel crossover operator based on the shortest edit path, aiming to overcome these challenges. 

\subsection{Network Encodings}
Architecture encodings in NAS refer to the representation of neural network architectures in a format that can be manipulated and searched efficiently. Numerous studies have explored diverse approaches to encode and represent neural architectures efficiently. Pioneering efforts, such as \cite{zoph2017neural}, introduced graph-based encodings where neural architectures are represented as directed acyclic graphs (DAGs). Nodes in the graph correspond to operations, while edges denote connections between them. This methodology allows for a flexible exploration of architecture space. The concept of breaking down architectures into repeatable cells, initially proposed by \cite{pham2018efficient}, has gained prominence. In this approach, architectures are constructed by stacking and connecting these cells, offering a modular and scalable representation. The utilization of continuous vectors or embeddings to represent architectures in a high-dimensional space has been proposed as an effective encoding strategy. This allows for efficient optimization through gradient-based methods. Learned vector-based auto-encoders have also shown promise in Arch2Vec \cite{arch2vec} which learns the compressed latent vector for encoding a neural architecture. Building on that work, CATE \cite{cate} utilizes transformers to perform computational-aware clustering of architectures. FLAN \cite{flan} proposes a flow attentive hybrid encoder that utilizes dense graph flow to counteract the over smoothing in GCNs. These prior methods provide a substantial comparison benchmark across a diverse set of tasks and search spaces.

\subsection{Search Space}
Neural Architecture Search (NAS) is intricately tied to the definition and exploration of search spaces, influencing the efficiency and effectiveness of automated architecture design. This section reviews key works that have contributed to the conceptualization and characterization of search spaces in the context of NAS.

Graph-based Search Spaces: Pioneering research, such as the work by \cite{real2017largescale}, introduced graph-based search spaces where neural architectures are modeled as directed acyclic graphs (DAGs). Nodes represent operations, and edges denote connectivity. This approach allows for a comprehensive exploration of architectural configurations, facilitating the discovery of intricate dependencies.

Cell-based Search Spaces: \cite{pham2018efficient} introduced the concept of cell-based search spaces, where architectures are constructed by repeating and stacking modular cells. These cells encapsulate architectural patterns, enabling a more structured and efficient exploration of the design space.

Regularized Search Spaces: Recent efforts, as seen in \cite{tan2018mnas}, have introduced regularization techniques to constrain the search space, promoting the discovery of architectures with desirable properties. This regularization aids in mitigating the vastness of the search space, enhancing the efficiency of NAS algorithms.

This survey highlights the evolution of search space definitions in NAS, showcasing a spectrum of approaches from graph-based representations to operation-level and hierarchical structures. The nuanced exploration of search spaces is crucial for advancing the state-of-the-art in automated neural architecture design.


\subsection{Zero-Cost NAS}
In traditional NAS methods, models are typically trained on a dataset to evaluate their performance and guide the search process. However, in zero-shot NAS, the search process doesn't rely on labeled data during the architecture search phase. Instead, it leverages surrogate objectives or proxy tasks by passing randomized inputs through an architecture with randomized weights. Through this process, zero-shot NAS is able to capture the expressivity of a networks which in turn has a high degree of correlation to the trained accuracy of the networks. This approach enables the discovery of neural network architectures that are effective for a wide range of tasks without the need for task-specific data. However, this approach comes with the draw back of reduced accuracy when compared with task-specific and traditional one-shot NAS. 

\section{Theoretical Analysis} 
\label{sec:math}
Through careful analysis we can show the precise reduction in the architecture space through difference projection. We begin by defining some constants. Let $n$ represents the number of nodes or depth of a neural architecture. Similarly, let $r$ represent the number of operation choices and $\mathcal{A}$ represent the set of candidate architectures. We can also define the difference of architecture set, $\mathcal{D}^k$, to represent the subtraction of 2 candidate architectures that are $k$ edits away from each other. 
\begin{definition}
$n$ is the number of nodes or depth of a network, $r$ is the number of operation choices, $\mathcal{A}$  is the set of candidate architectures being considered by NAS.
\end{definition}
There are two basic search space types in NAS: block-based search spaces and cell-based search spaces. Block-based search spaces are combinatorial search spaces where the depth of the network and connections in the network are predefined. NAS is only searching for the operation type at each node in the network. Hence, the magnitude of a block-based search space is defined as follows:
\begin{theorem}
    \label{th1}
    Given a block-based search space represented by a list of operations
    $\vert \mathcal{A} \vert = r^n$
\end{theorem}
In cell-based search spaces, NAS is responsible for finding the best way to connect nodes in a cell and what operations to place at each node in the cell. The cell is then stacked repeatedly to produce the final architecture. Cell-based search spaces are typically contain significantly less nodes than block-based search spaces but add complexity back through searching for the structure of the cell. Cell structure is typically represent as an adjacency matrix with the upper triangle containing information about which nodes are connected to which other nodes. The size of the upper triangle is $\frac{n(n-1)}{2}$. It follows that:
\begin{theorem}
    \label{th2}
    Given a cell-based search space represented by an adjacency matrix
    $\vert \mathcal{A} \vert = \frac{n(n-1)}{2} r^n$
\end{theorem}
Now that we have established the baseline size of NAS search spaces, we can begin to examine the impact of difference projection. In an unconstrained setting, difference projection does not reduce the size of the search space. Taking a difference of two random architectures results in a difference matrix of the same cardinally. 
\begin{theorem}
    \label{th3}
    Difference of architectures generates and equally high dimensional space
    $\vert \mathcal{A} \vert \approx \vert \mathcal{D} \vert$
\end{theorem}
\begin{proof}
For $a_i,a_j \in \mathcal{A}$ \\
Let $a_i - a_j = d_x$ and $d_x \in \mathcal{D}$ \\
$\vert \mathcal{A} \vert \approx \vert \mathcal{D} \vert$
\end{proof}
Simply taking the difference doesn't help to achieve reduction in size. In order to obtain reduction in size, we must constrain the networks that we are taking the difference of. If we limit the networks to be identical, taking their difference would result in nothing. This provides no value as we no longer have any information to predict on. Let's suppose we limit the networks to be "close". The networks can be identical except for one change. Now, when we take the difference, there is precisely one value in the difference representation as opposed to the architecture representation that has a significantly larger state representation. The single change can take place at one of $n$ nodes and take one of $r$ values, hence: 
\begin{theorem}
    $ \vert \mathcal{D}^1 \vert = rn$
\end{theorem}
More formally, we have:
\begin{proof}
For $a_i \in \mathcal{A}$ let $a_i^1$ represent architecture that is one edit away from $a_i$.\\
Let $a_i - a_i^1 = d_i^1$ and $d_i^1 \in \mathcal{D}^1$
$ \vert \mathcal{D}^1 \vert = rn$ \\
$\vert \mathcal{A} \vert \gg \vert \mathcal{D}^1 \vert$
\end{proof}
\begin{corollary}
    Difference of close architectures generates a lower dimensional space
    $\vert \mathcal{A} \vert \gg \vert \mathcal{D}^1 \vert$
\end{corollary}
While this statement holds in the $\mathcal{D}^1$ space, a more generic formula for size in different difference spaces would be nice to have. This flexibility allows for larger difference spaces with more information. Suppose we have a difference space with $k$ changes between the two architectures being subtracted. What would the size of this space be? We can show:
\begin{theorem}
$\vert \mathcal{D}^k \vert = r^k \binom{n}{k} $
\end{theorem}
The full proof by induction can be seen below. 
Now we have a full spectrum of difference of architecture spaces we can use, each with a different size. We have also shown the baseline size of NAS search spaces along with the size of the space resulting from the difference projection. Difference of architecture projection eliminates the exponential growth with respect to depth allowing for a greater number of nodes in both cell-based and block-based NAS search spaces when looking at close neighbor. This particular trait comes in handy when performing NAS on large spaces as we will elaborate in the next section.   
\begin{proof}
\label{k-delta}
\begin{align*}
Base case: \vert \mathcal{D}^1 \vert &= rn \\
\vert \mathcal{D}^2 \vert &= r^2\frac{n(n-1)}{2} \\
\vert \mathcal{D}^3 \vert &= r^3\frac{n(n-1)(n-2)}{6} \\
Assume: \vert \mathcal{D}^k \vert &= r^k\frac{n!}{k!(n-k)!} \\
 &= r^k {n \choose k} \\
Show: \vert \mathcal{D}^{k+1} \vert &= r^{k+1}\frac{n!}{(k+1)!(n-(k+1))!} \\
&= \left(r^k\frac{n!}{k!(n-k-1)!}\right)\frac{r}{k+1} \\
&= \left(r^k\frac{n!}{k!(n-k-1)!}\right)\frac{r(n-k)}{(k+1)(n-k)} \\
&= \left(r^k\frac{n!}{k!(n-k)!}\right)\frac{r(n-k)}{(k+1)} \\
&= \vert \mathcal{D}^k \vert \frac{r(n-k)}{(k+1)} \\
Given:  \vert \mathcal{D}^k \vert &= r^k\frac{n!}{k!(n-k)!} \\
\vert \mathcal{D}^k \vert \frac{r(n-k)}{(k+1)} &= \left(r^k\frac{n!}{k!(n-k)!}\right)\frac{r(n-k)}{(k+1)} \\
\vert \mathcal{D}^{k+1} \vert &= r^{k+1}\frac{n!}{(k+1)!(n-(k+1))!}
\qedhere
\end{align*}
\end{proof}

\section{Methods}
\label{sec:methods}
\begin{figure*}[t!]
    \includegraphics[width=.99\linewidth]{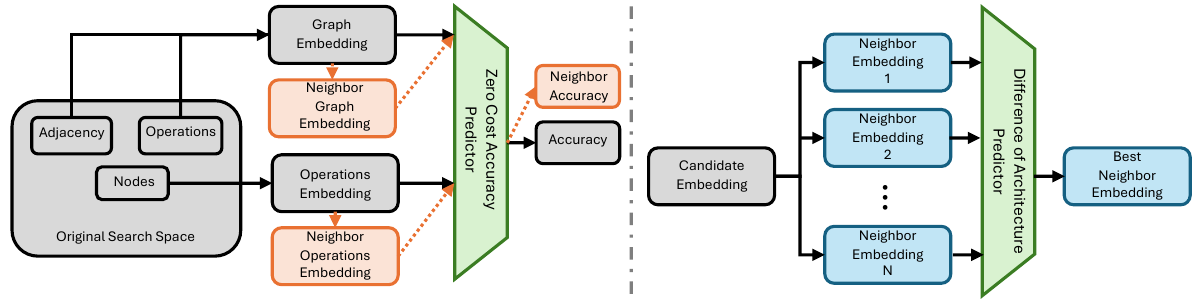}
    \caption{The left side of the figure shows the process for generating the difference of architecture dataset. Both pairs of graph and operation embeddings along with both accuracies are stored to create the DoA dataset. The right side of the figure shows the DoA predictor used to generate neighbors during the modified evolutionary search.}
    \label{fig:detail}
\end{figure*}

\begin{algorithm}[b]
\caption{Generate Difference of Architecture Dataset}\label{alg:datadelta}
\begin{algorithmic}
\State $Z \gets \texttt{empty set}$
\For{\texttt{number of desired samples}}
        \State $a \gets A$
        \State $d \gets D^{1}(a)$
        \State $j \gets \texttt{DeltaArch}(a,d)$ \Comment{Encoding of difference between $a$ and $d$}
        \State $k \gets \texttt{DeltaAcc}(a,d)$ \Comment{Difference in zero-shot accuracy between $a$ and $d$}
        \State $Z \gets (j,k) $
\EndFor
\State \textbf{return} Z
\end{algorithmic}
\end{algorithm}

 The difference projection step is able to able to reduce the size of the search space. However, we are still presented with two main problems. Firstly, how do we obtain the difference of architecture space? Secondly, how do we perform NAS in the difference of architecture space?

\subsection{Difference of Architectures}
To answer our first question, we begin by sampling pairs of networks and computing the difference in predicted accuracy. In order to construct the difference of architecture space, we must first obtain samples of pairs of networks that are close in edit distance. To achieve this, we randomly sample networks across the search space then perform a series of changes to obtain neighboring networks at a given edit distance. These pairs of networks are used to construct difference of architecture space $D^k$ and have the varying carnality as specified above. Importantly, the difference of architecture encoding is a many-to-one encoding meaning that we can obtain several pairs of networks with the same difference. The difference in accuracy is obtained using a zero-cost proxy initially proposed by ZenNAS\cite{zeroshot}. While the zero-cost proxy trades accuracy for computational cost, we are able to improve the accuracy in the difference space through obtaining multiple samples for the same encoding. Through this technique, we are able to boost the accuracy of the zero-cost proxy and the difference predictor. We store the difference pair and value in the a custom dataset that will be then used to train the difference predictor. Note that this method has 2 hyperparameters to tune: the $k$ in $D^k$ and the number of samples for each difference encoding $n$. Through experimentation, we found that $k=1$ and $n=4$ provide sufficient results. The full results from the hyperparameter sweep can be found in the appendix. Algorithm \ref{alg:datadelta} demonstrates the technique for building a difference of architecture space. The result is a dataset consisting of pairs of networks and their difference in accuracy. This dataset is then used to construct a predictor that given a pair of networks predicts the difference in accuracy of those networks. Since this predictor is trained for a specific $k$, we limit the inputs to the predictor to also be constrained by the same $k$. 
That is to say, if we are training a predictor for the $D_1$ space, that predictor can be queried for networks that are at most one edit away from each other. 

\subsection{NAS algorithm}
\begin{algorithm}[b]
\caption{Delta-NAS}\label{alg:delta}
\begin{algorithmic}
\State $F(a_i, a_j)$ \Comment{Build a predictor using DoA dataset}
\State $As$ \Comment{Initialize population randomly}
\While{Convergence condition not met}
    \For{$\texttt{p} \in \texttt{As}$}
        \For{$\textit{pn} \in D^{1} \texttt{of} p$}
            \State $mp \gets max(F(p, pn))$ \Comment{Keep best neighbor}
        \EndFor
        \State $An \gets mp$ \Comment{Store best neighbor of p }
    \EndFor
    \State $As \gets An$
\EndWhile
\State $\textbf{return } max(As)$

\end{algorithmic}
\end{algorithm}
Now that we have projected the problem of NAS to a lower dimensional space, we are faced with another problem: How do we use the difference predictor to perform NAS? Traditionally, the final step of the NAS pipeline is using the predictor to query for the accuracy of many candidate networks and perform some search algorithm. Most commonly this search algorithm is derived from an evolutionary method where a population of networks is selected and repeatedly refined and resampled producing a more fit population in progressive steps until finally we can sift through the remaining networks and pick the one with the highest performance. This approach breaks down initially when using a difference predictor as we no longer have the raw accuracy of a given candidate network. Instead, we must derive a modified approach. Similar to evolutionary methods we begin by sampling a population of networks. Then for each network, we query the difference predictor for the most promising neighbor through sampling single edit changes. We then take a step in this direction. This process is repeated for a sufficient number of networks and iterations until convergence is reached. Sampling many initial networks and continuing to resample during the evolutionary algorithm helps avoid the issue of getting stuck in local minima. As with all NAS techniques, there is a trade-off between the number of networks queried and the probability of getting stuck in a non-optimal solution. By using the difference predictor instead of a traditional predictor, the evolutionary algorithm is able to converge significantly faster as shown in Figure \ref{fig:con-nb-ged}. We employ evolutionary search following many of the hyperparameters selected by SEP \cite{pmlr-v202-qiu23b-sep} and further outlined in the Appendix \ref{evohyp}.



\section{Experimental Results}
\label{sec:exp}
\begin{table*}
\centering
\caption{Comparison of accuracy prediction when using different encoding methods. Percentage of search space used in training predictors varies between 1 and 10 percent.}
\label{table:acc}
\begin{tabular}{l|cccccccc}
Search Space        & \multicolumn{2}{c}{\begin{tabular}[c]{@{}c@{}}NASBench101\\ (\% of 423k nets)\end{tabular}} & \multicolumn{2}{c}{\begin{tabular}[c]{@{}c@{}}NASBench201\\ (\% of 15.6k nets)\end{tabular}} & \multicolumn{2}{c}{\begin{tabular}[c]{@{}c@{}}TransNASBench\\ (\% of 7k nets)\end{tabular}} & \multicolumn{2}{c}{\begin{tabular}[c]{@{}c@{}}NASBench301\\ (\% of 9k samples)\end{tabular}} \\ \hline
Encoding            & 1\%                                          & 10\%                                         & 1\%                                           & 10\%                                         & 1\%                                          & 10\%                                         & 1\%                                           & 10\%                                         \\ \hline
ADJ                 & 0.327                                        & 0.514                                        & 0.382                                         & 0.559                                        & 0.331                                        & 0.654                                        & 0.401                                         & 0.560                                        \\
Path                & 0.387                                        & 0.752                                        & 0.396                                         & 0.781                                        & 0.350                                        & 0.756                                        & 0.343                                         & 0.776                                        \\ \hline
ZCP                 & 0.591                                        & 0.684                                        & 0.376                                         & 0.638                                        & 0.545                                        & 0.721                                        & 0.272                                         & 0.691                                        \\ \hline
Arch2Vec            & 0.210                                        & 0.345                                        & 0.144                                         & 0.356                                        & 0.298                                        & 0.420                                        & 0.228                                         & 0.416                                        \\
CATE                & 0.632                                        & 0.467                                        & 0.571                                         & 0.477                                        & 0.682                                        & 0.701                                        & 0.349                                         & 0.675                                        \\ \hline
FLAN                & 0.665                                        & 0.823                                        & 0.782                                         & 0.812                                        & 0.632                                        & 0.837                                        & 0.537                                         & 0.810                                        \\ \hline
\textbf{DoA (ours)} & \textbf{0.701}                               & \textbf{0.852}                               & \textbf{0.759}                                & \textbf{0.832}                               & \textbf{0.692}                               & \textbf{0.845}                               & \textbf{0.795}                                & \textbf{0.831}                              
\end{tabular}
\end{table*}
\subsection{Setup}
\subsubsection{Datasets and Search Spaces}
We conduct extensive experiments on the popular image recognition datasets (CIFAR10\cite{krizhevsky2009learning}, CIFAR100\cite{krizhevsky2009learning}, and ImageNet\cite{imagenet}) in addition to several other image related tasks. For object and scene classification we use the Taskonomy\cite{zamir2018taskonomy} and Places\cite{places} datasets respectively. Finally for semantic segmentation, we utilize the MSCOCO dataset\cite{lin2015microsoft}.

To perform NAS, we use both cell-based and block-based search spaces. NASLib\cite{naslib-2020} is a modular and flexible framework wrapping many common NAS search spaces. We utilize NASBench101\cite{ying2019nasbench} which contains 432k unique architectures using a backbone of cells stacked in a repeated fashion. Each cell contains 5 nodes and up to 7 edges and 1 of 3 operations is placed in each node. In NASBench201\cite{dong2020nasbench201}, architecture are comprised of 4 nodes and 5 operations resulting in 15k networks that are repeatedly trained providing a high quality training curve for each architecture. TransNASBench\cite{transNASBench} evaluates each of it's 7k networks across 7 tasks resulting in 50k network/accuracy pairs. NASBench301\cite{zela2022surrogatenasbenchmarksgoing} is the largest space we consider with $10^{21}$ architectures. This block-based search space contains architectures represented as a stack of operations with a depth of 20 and each operation can be one of 4 potential operations.  
\subsubsection{Implementation Details}
\begin{figure*}[]
  \centering
    \begin{subfigure}{0.32\linewidth}
    \includegraphics[width=.9\linewidth]{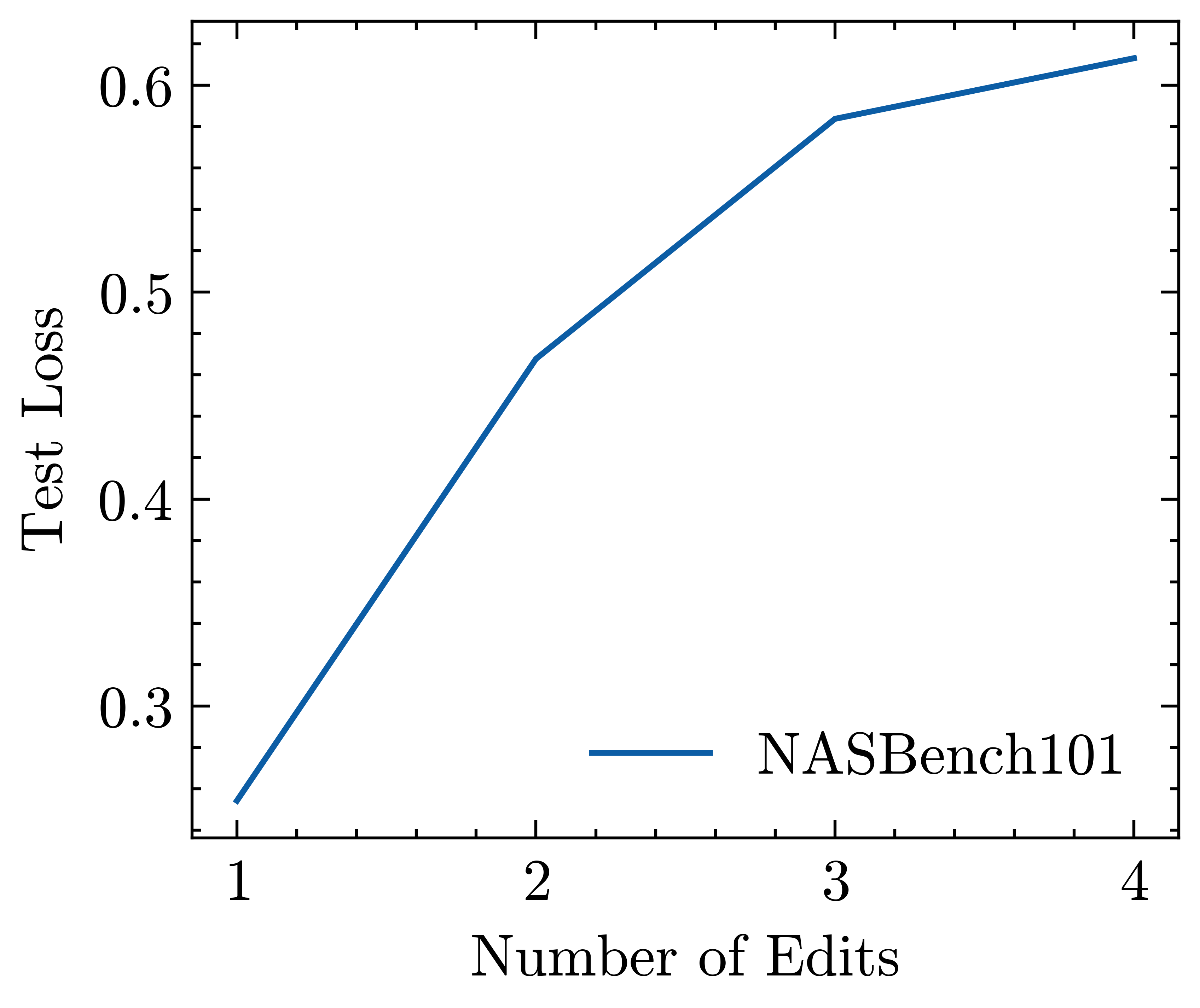}
    \caption{Predictor loss vs edit distance in NASBench101}
    \label{fig:loss-nb1}
  \end{subfigure}
  \hfill
  \begin{subfigure}{0.32\linewidth}
    \includegraphics[width=.9\linewidth]{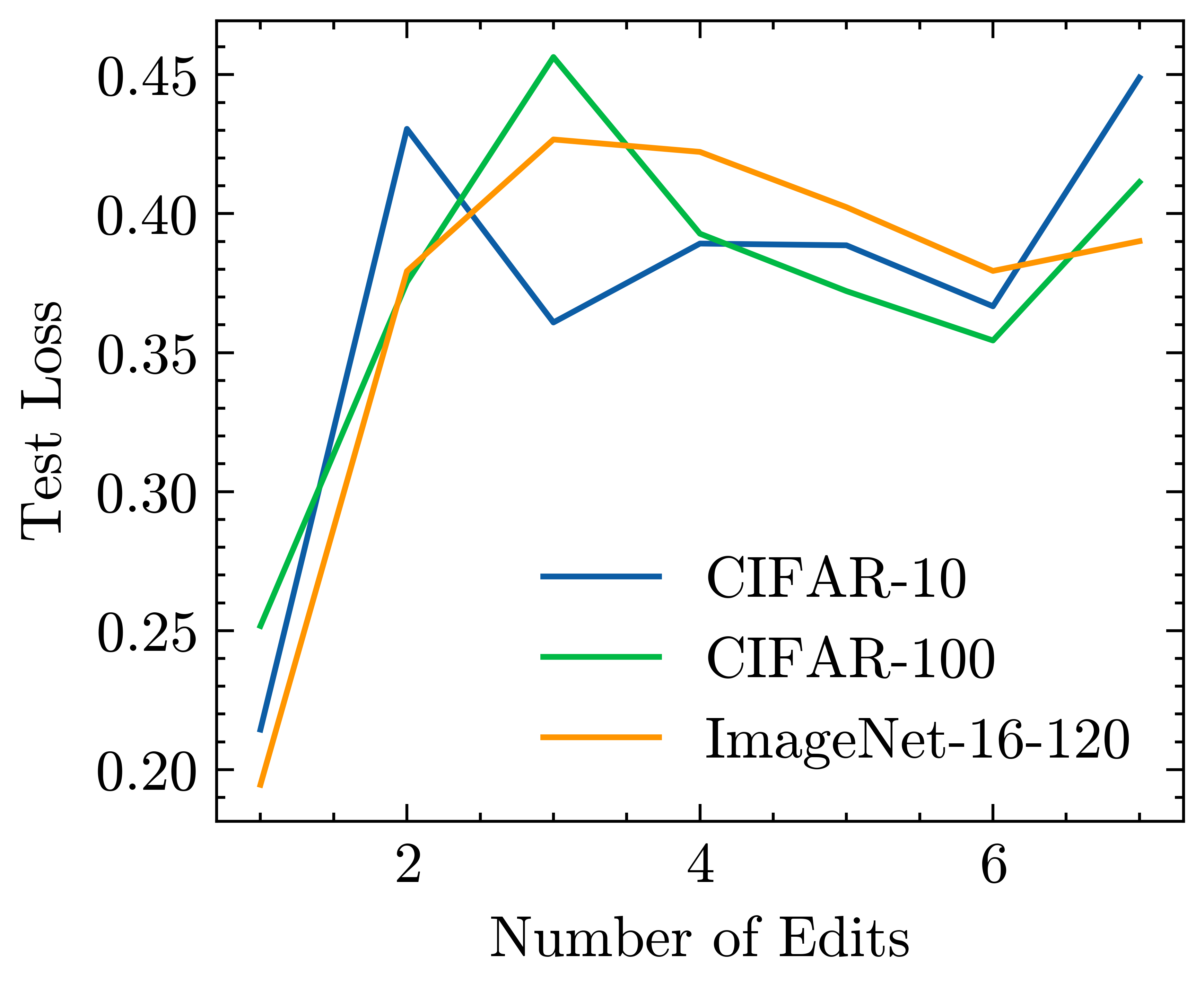}
    \caption{Predictor loss vs edit distance in NASBench201}
    \label{fig:loss-nb2}
  \end{subfigure}
  \hfill
  \begin{subfigure}{0.32\linewidth}
     \includegraphics[width=.9\linewidth]{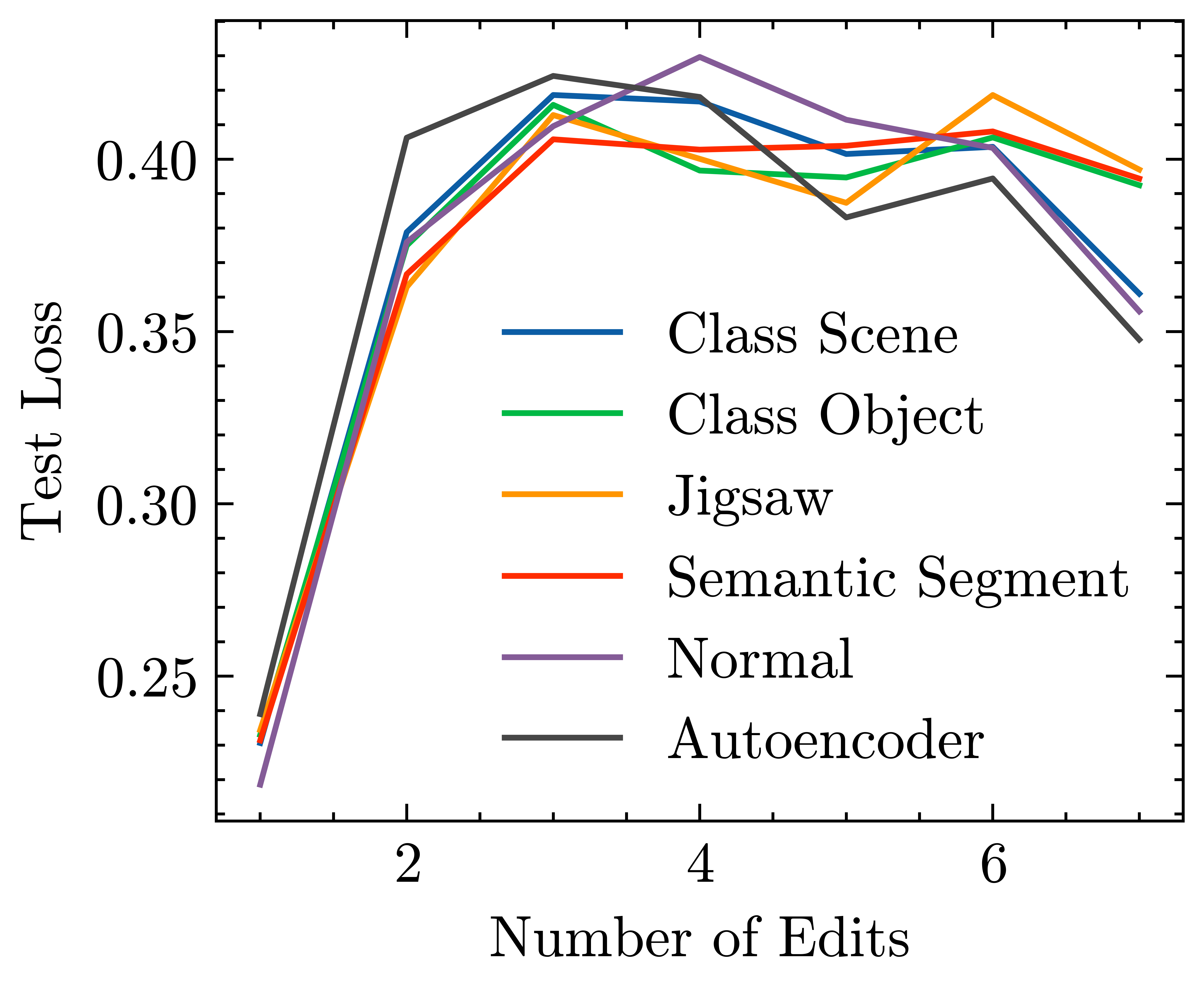}
    \caption{Predictor loss vs edit distance in TransNASBench}
    \label{fig:loss-tnb}
  \end{subfigure}
  \caption{Predictor loss increases greatly as number of edits increases beyond 1 and plateaus.}
  \label{fig:short}
\end{figure*}
We first construct a difference of architectures space with varying $k$ for each search space, NASBench101, NASBench201 and TransNASBench using Algorithm \ref{alg:datadelta}. Next, we train a predictor on each DoA dataset and plot the results in Figure \ref{fig:short}. From this, we observe that the predictor loss increases significantly when $k\geq 2$. This phenomenon is more pronounced in NASBench201 and TransNASBench likely caused by the increased graph complexity of these search spaces. The trend in NASBench101 is fairly linearly however the best predictor performance still occurs with $k=1$. More importantly, $k=1$ results in the greatest reduction in search space size. Only considering imeditate neighbors for each candidate also results in the smallest cardinality of the DoA space. Hence, we chose to move forward with $D_1$ as the default difference of architecture space. Next, we then perform NAS using a predictor trained on this space and the modified evolutionary algorithm \ref{alg:delta}. Please see Appendix \ref{evohyp} for hyperparammeter details for the modified evolutionary algorithm on each of the search spaces.

\subsection{Predictor Results}


Across all search spaces and tasks, any number of edits greater than one causes a significant decline in accuracy in the difference predictor. The many-to-one nature of the difference encoding likely causes the degradation of performance as stacking these changes greatly increases the variability of the accuracy difference between the pair of networks. These findings generalize well across a diverse range of search space sizes, tasks, and datasets. Using a $D^1$ space proves to be the best choice for projection. Importantly, this projection also reduces the size of the search space most significantly and contains the fewest encoded elements. Another important benefit of the many-to-one difference encoding is the reduction in graph isomorphism. Since the individual structure of the architectures themselves are removed the new latent space, there are no longer isomorphisms.

In the initial phase of NAS, we develop a difference predictor by leveraging the difference in architecture projections. This approach, based on difference encoding, proves to be highly effective, as evidenced by the results in Table \ref{table:acc}, where it consistently outperforms existing encoding schemes across various scenarios. Specifically, we sample 1\% and 10\% of the total networks within each search space. These subsets are subsequently used to train the performance predictor according to the methodology associated with each encoding scheme. To ensure robustness, the predictor's performance is evaluated on a separate hold-out set, with the Kendall’s Tau correlation coefficient reported as the evaluation metric, as shown in Table \ref{table:acc}.

We further extend these experiments across three distinct search spaces, encompassing a total of nine individual benchmarks, providing a comprehensive evaluation of our method's effectiveness. For TransNASBench in particular, we aggregate results across multiple tasks, ensuring equal weighting to maintain fairness in performance comparison. This thorough experimentation highlights the generalizability and robustness of our difference encoding technique, as it consistently demonstrates superior prediction accuracy and efficiency across a diverse range of NAS benchmarks and search spaces.Complete results for each task in TransNASBench are available in Appendix \ref{tnb-extend}.




\subsection{NAS Results}

\begin{figure*}
  \centering
    \begin{subfigure}{0.47\linewidth}
    \includegraphics[width=.9\linewidth]{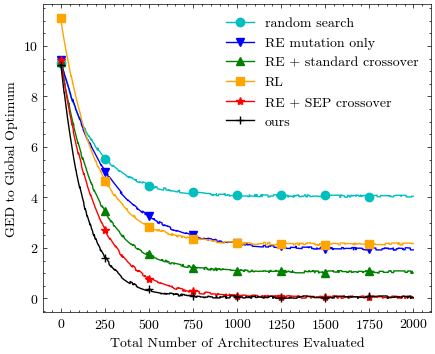}
    \caption{Convergence Plot on NASBench101 (50 runs)}
    \label{fig:con-nb-ged}
  \end{subfigure}
  \hfill
  \begin{subfigure}{0.47\linewidth}
    \includegraphics[width=.9\linewidth]{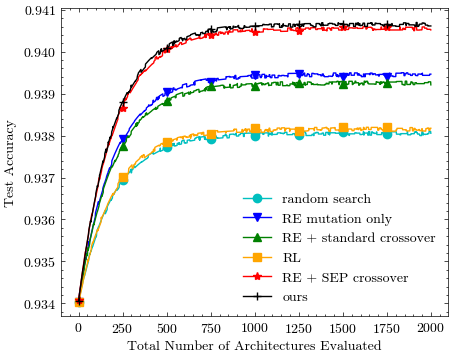}
    \caption{Convergence Plot on NASBench101 (50 runs)}
    \label{fig:con-nb-acc}
  \end{subfigure}
  \caption{Delta-NAS converges significantly faster than existing encoding schemes and evolutionary based methods.}
  \label{fig:short1}
\end{figure*}

The use of difference-based architecture encoding significantly enhances sampling efficiency compared to traditional encoding and search methodologies. As demonstrated in Figure \ref{fig:con-nb-ged}, our encoding and search technique achieves rapid convergence towards the optimal network architecture. The y-axis in this figure illustrates the average edit distance to the optimal design in NASBench101, plotted against the number of samples. This metric clearly shows the superior efficiency of our approach in navigating the search space. Furthermore, Figure \ref{fig:con-nb-acc} highlights the test accuracy as a function of the number of networks evaluated, further confirming the effectiveness of our method.

\begin{figure*}
  \centering
    \begin{subfigure}{0.47\linewidth}
    \includegraphics[width=.9\linewidth]{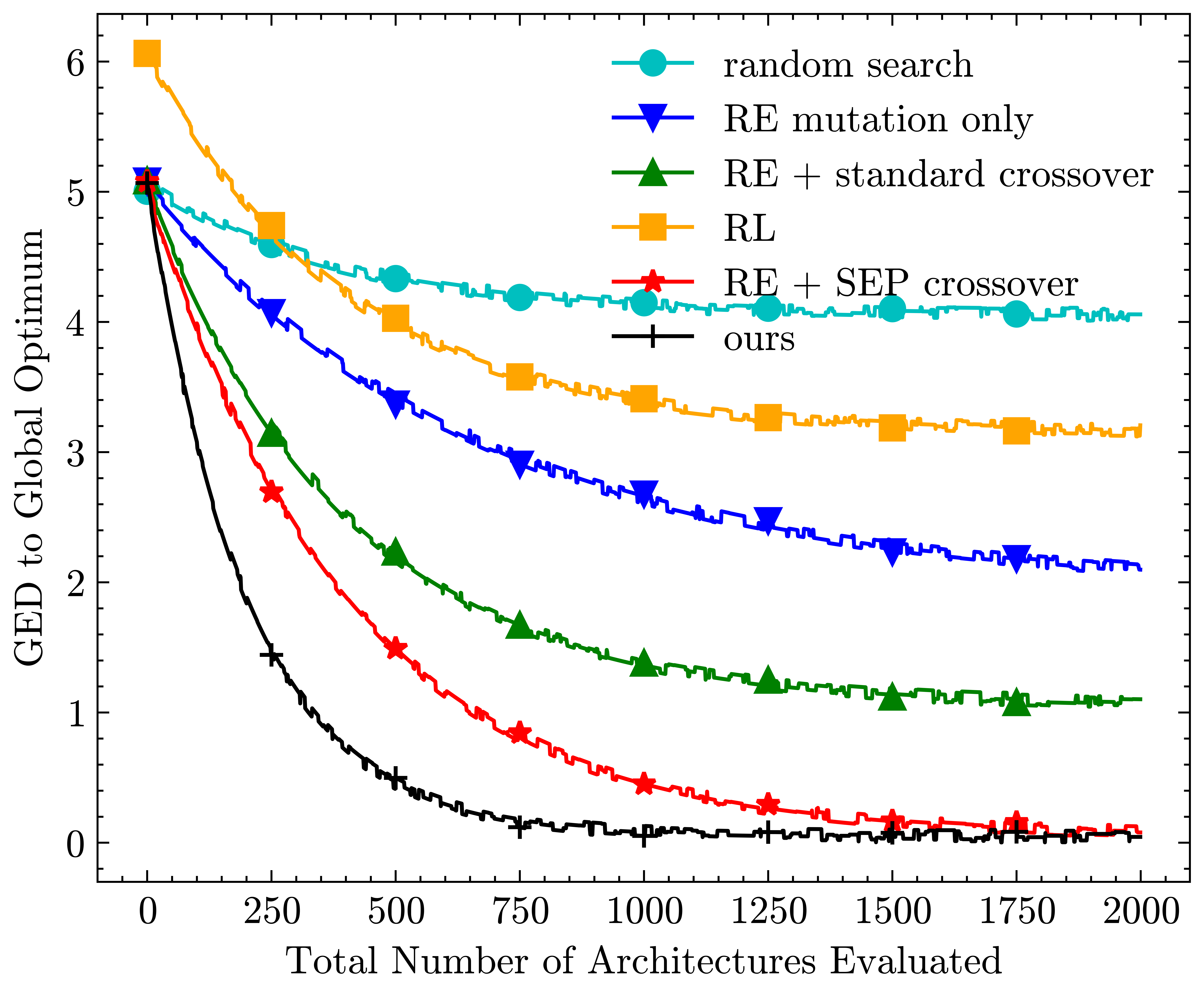}
    \caption{Convergence Plot on NASBench201 (50 runs)}
    \label{fig:con-nb201-ged}
  \end{subfigure}
  \hfill
  \begin{subfigure}{0.47\linewidth}
    \includegraphics[width=.9\linewidth]{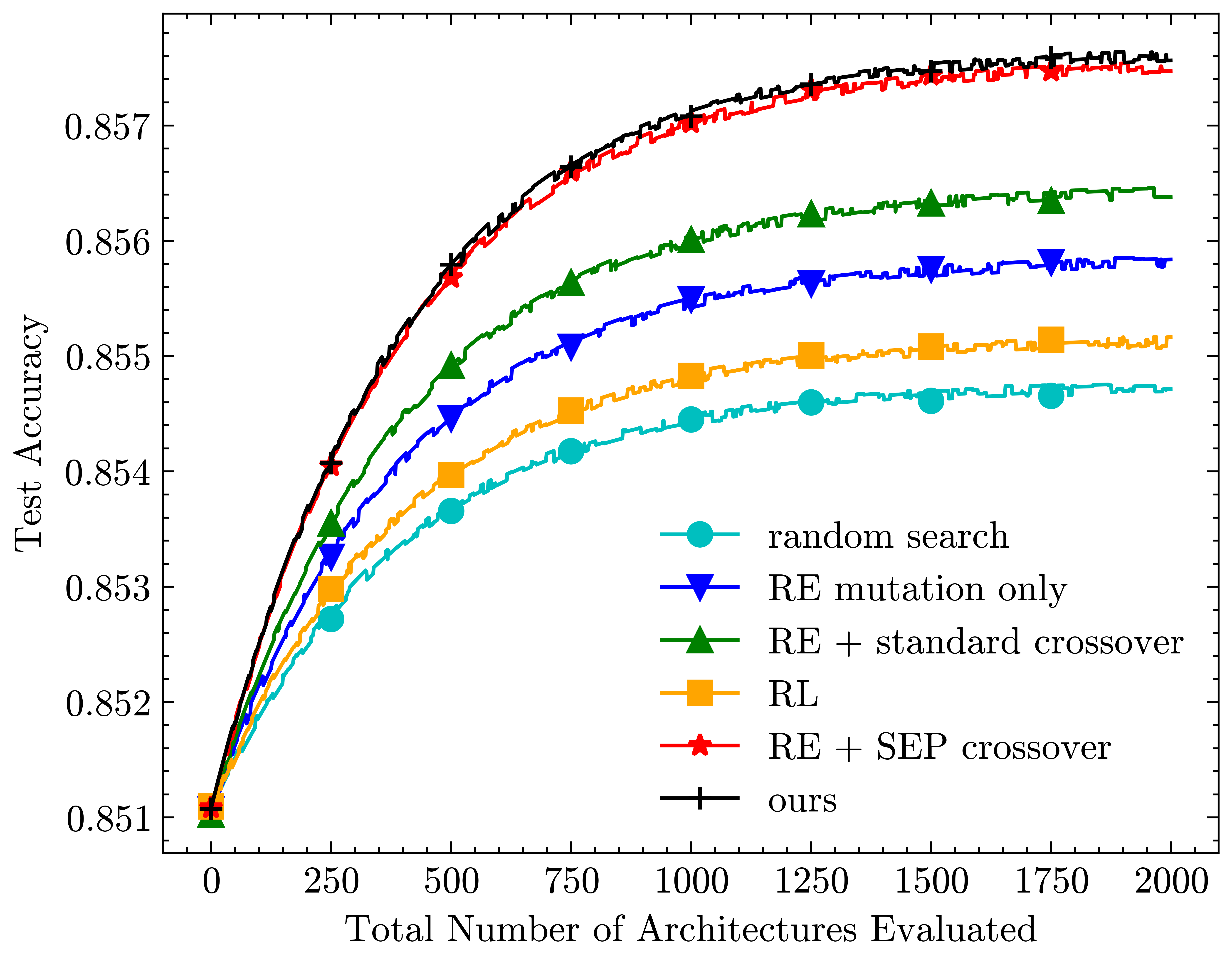}
    \caption{Convergence Plot on NASBench201 (50 runs)}
    \label{fig:con-nb201-acc}
  \end{subfigure}
  \caption{In larger search spaces, the difference in performance between Delta-NAS and previous works is more pronounced. }
  \label{fig:short2}
\end{figure*}

Our approach surpasses the current state-of-the-art technique, regularized evolution with shortest edit path crossover \cite{pmlr-v202-qiu23b-sep}, in terms of both convergence speed and final performance. In addition to outperforming this method, we also demonstrate substantial improvements over other established techniques, including regularized evolution with standard crossover, regularized evolution alone, random search, and reinforcement learning-based search methods. This performance superiority is consistent across both NASBench101 and NASBench201, as depicted in Figure \ref{fig:con-nb201-acc}, where our method consistently achieves higher accuracy and efficiency. Importantly, our technique performs better across the board in this search space as well 

A critical aspect of our evaluation is the rigorous comparison with baseline methods. To ensure a fair and unbiased comparison, we maintain all hyperparameters constant between the regularized evolution baseline and our modified version, thus isolating the impact of our difference encoding and shortest edit path crossover. These experimental results underline the robustness and effectiveness of our approach across multiple benchmarks, making it a compelling advancement over existing NAS techniques.

\section{Conclusion}
\label{sec:conc}
Neural Architecture Search (NAS) remains pivotal in the design and development of task-specific neural networks, but modern NAS techniques face challenges related to increasing search space complexity and computational costs. Current approaches fall into two categories: fine-grained but computationally expensive NAS, and coarse-grained, low-cost NAS. Our work aimed to bridge this gap by creating an algorithm capable of fine-grained NAS at a low cost. By projecting the problem into a lower-dimensional space through predicting the accuracy difference between pairs of similar networks, we reduced computational complexity from exponential to linear in relation to the search space size. Our algorithm is backed by a strong mathematical foundation and extensive experimental validation across multiple NAS benchmarks. The results demonstrate that our method significantly outperforms existing approaches, achieving better performance with much higher sample efficiency.

{\small
\bibliographystyle{ieee_fullname}
\bibliography{delta-nas-wacv}
}

\clearpage
\section{Appendix}
\label{sec:apen}

\subsection{Experimental Setup Details}
\label{evohyp}
We follow standard NAS practices when using benchmark datasets. Training info and creation of the benchmark datasets can be found in the respective publications \cite{ying2019nasbench,dong2020nasbench201,transNASBench,zela2022surrogatenasbenchmarksgoing}. Default common settings for image transformations are used. Namely mix-up, label-smoothing, random crop, resize, flip and AutoAugment \cite{cubuk2019autoaugmentlearningaugmentationpolicies}. We also follow stand convention in evolutionary settings with a population size of 256, number of evolutionary iteration as $T=96k$.  



\newpage
\newpage

\end{document}